\documentclass[conference]{IEEEtran}
\IEEEoverridecommandlockouts
\usepackage{cite}
\usepackage{amsmath,amssymb,amsfonts}
\usepackage{algorithmic}
\usepackage{graphicx}
\usepackage{textcomp}
\usepackage{xcolor}
\usepackage{comment}
\usepackage{tcolorbox}
\usepackage{diagbox} 
\usepackage{colortbl}
\usepackage{color}
\usepackage{booktabs}
\usepackage{multirow}
\usepackage{hyperref}
\definecolor{SpeechColor}{RGB}{197, 90, 17}
\definecolor{TextColor}{RGB}{47, 85, 151}

\def\BibTeX{{\rm B\kern-.05em{\sc i\kern-.025em b}\kern-.08em
    T\kern-.1667em\lower.7ex\hbox{E}\kern-.125emX}}
    
\begin{document}

\title{SpeakStream: Streaming Text-to-Speech with Interleaved Data}

\author{\IEEEauthorblockN{1\textsuperscript{st} Richard He Bai}
\IEEEauthorblockA{\textit{Apple} \\
United States \\
richardbai@apple.com}
\and
\IEEEauthorblockN{2\textsuperscript{nd} Zijin Gu}
\IEEEauthorblockA{\textit{Apple} \\
United States \\
zijin@apple.com}
\and
\IEEEauthorblockN{3\textsuperscript{rd} Tatiana Likhomanenko}
\IEEEauthorblockA{\textit{Apple} \\
United States \\
antares@apple.com}
\and
\IEEEauthorblockN{4\textsuperscript{th} Navdeep Jaitly}
\IEEEauthorblockA{\textit{Apple} \\
United States \\
njaitly@apple.com}
}

\maketitle

\begin{abstract}

The latency bottleneck of traditional text-to-speech (TTS) systems fundamentally hinders the potential of streaming large language models (LLMs) in conversational AI. 
These TTS systems, typically trained and inferenced on complete utterances, introduce unacceptable delays – even with optimized inference speeds – when coupled with streaming LLM outputs. 
This is particularly problematic for creating responsive conversational agents where low first-token latency is critical.
In this paper, we present SpeakStream, a streaming TTS system that generates audio incrementally from streaming text using a decoder-only architecture. 
SpeakStream is trained using a next-step prediction loss on interleaved text-speech data. 
During inference, it generates speech incrementally while absorbing streaming input text, making it particularly suitable for cascaded conversational AI agents where an LLM streams text to a TTS system. 
Our experiments demonstrate that SpeakStream achieves state-of-the-art latency results in terms of first-token latency while maintaining the quality of non-streaming TTS systems.
Our demo website is available at \url{https://apple.github.io/speakstream-demo}.

\end{abstract}


\section{Introduction}
Recent years have witnessed a surge of interest in speech interfaces for large language models (LLMs). 
While substantial research has focused on end-to-end models where LLMs directly generate tokenized audio~\cite{borsos2023audiolm, defossez2024moshi, Qwen2.5-Omni}, studies indicate that cascaded models—which stream text from LLMs to text-to-speech (TTS) systems—consistently outperform end-to-end approaches~\cite{nguyen2025spirit,sakshi2024mmau}.

A primary challenge in cascaded models is reducing latency, which stems from two main sources: (1) waiting for the LLM to generate a complete text segment (e.g., sentence) and (2) waiting for the TTS system to generate audio. 
To address text-waiting latency, recent approaches~\cite{dang2024zero,dekel2024speak} have explored partial text context windows to balance responsiveness and quality. 
However, these streaming methods often struggle with long-range dependencies and require careful tuning of text-speech alignment~\cite{dang2024zero,yang2024interleaved}. 
Furthermore, the separation between text encoding and speech generation processes can lead to suboptimal context utilization, particularly when modeling prosody across sentence boundaries.

However, a more fundamental source of latency arises from the architecture of traditional TTS systems themselves. 
These systems~\cite{bai2024dmel,pmlr-v162-bai22d,casanova2022yourtts,du2024cosyvoice,gao2023e3,ren2020fastspeech,wang2017tacotron,openai_tts} are designed to process complete text segments before generating any audio, introducing significant delays even with optimized inference speeds.  
While efforts like StreamSpeech~\cite{10096566} focus on accelerating the processing of complete utterances, they do not address the inherent latency caused by waiting for the entire text segment to arrive. 
This limitation is particularly problematic in interactive applications where minimizing the "first-token latency"---the time between receiving the first text token and hearing the first audio token---is crucial for a natural and responsive user experience.

\begin{figure}
    \centering
    \includegraphics[width=0.75\linewidth]{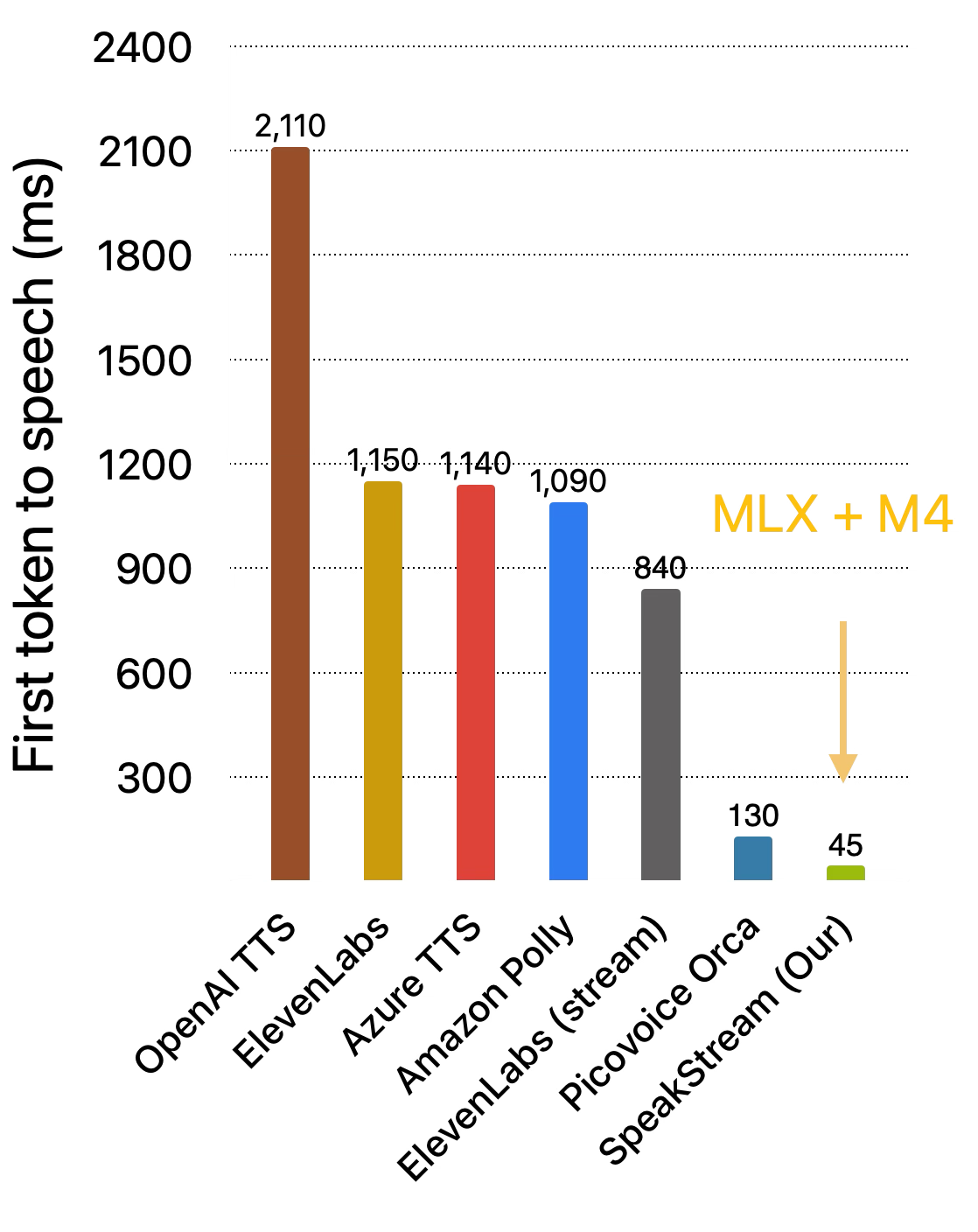}
    \caption{First-token to speech (time taken from the moment the LLM produces the first text token, until the TTS produces the first byte of speech) latency comparison between SpeakStream and other baselines. The 45ms latency is tested with MLX~\cite{mlx2023} and Mac Mini with M4 Pro chip and 64GB. Baselines latency is tested from \href{https://github.com/Picovoice/tts-latency-benchmark}{tts-latency-benchmark}.}
    \label{fig:latency}
    \vspace{-0.2cm}
\end{figure}
To overcome these limitations, we propose SpeakStream—a novel decoder-only TTS model that enables streaming TTS through modeling interleaved text-speech segments instead of a pair of text and speech. 
To create interleaved text-speech training data, we use a force-aligner~\cite{pmlr-v162-bai22d} to align text and speech pairs from traditional TTS datasets.
The model then trained like a standard LLM on the interleaved data, but the loss is computed only on the speech tokens and each speech segment is wrapped with a beginning-of-speech~(BOS) token and an end-of-speech~(EOS) token.
During inference, when the streaming text reaches enough length, a BOS token is appended after the last text token and the model will generate the corresponding speech segment.
After the EOS is generated, the model will take the upcoming text segment in, and repeat the process.
All the previous text and speech segments are cached in the key-value cache (kv-cache), so the model can reuse them for efficient inference.
Our approach offers three key advantages: (1) unified sequence modeling across modalities (a single decoder instead of separate encoders and decoders), (2) elimination of explicit alignment mechanisms during inference, and (3) efficient computation through kv-caching.
The transformer decoder learns to predict the next speech segment conditioned on the current text segment, previous speech segments, and previous text segments. This creates a coherent, unified context that captures both modalities, allowing the model to maintain complete information about previously generated speech while processing incoming text. 

We conduct extensive ablation studies to analyze the impact of text segment length and speech segment length. The ablation results show that it is important to provide additional text for each segment when doing interleaved training.
Also, the text repeated interleaving schema is better than without text repeats.
Our empirical results demonstrate SpeakStream's effectiveness: automatic evaluation shows it achieves the lowest error rate across all latency configurations, while human evaluators rate its coherence comparable to non-streamed systems like RichTTS~\cite{bai2024dmel}. 
By deploying SpeakStream to a Mac mini (M4 Pro chip, 64GB, 2024), we show it achieves 30ms TTS latency plus 15ms vocoder and player latency in Figure~\ref{fig:latency}, making it suitable for real-time interactive applications.
\section{Related Work}
Traditional TTS systems~\cite{wang2017tacotron,casanova2022yourtts,ren2020fastspeech,gao2023e3,du2024cosyvoice} process complete text to generate complete audio. However, the rise of conversational AI demands reduced latency through dual-streaming capabilities, i.e., streaming text input and streaming audio output. 

Dual-streaming TTS mimics how humans read aloud a text stream as it unfolds. Modeling this behavior with neural networks presents several challenges. First, when synthesizing speech streamingly, creating smooth transitions between audio chunks while avoiding artifacts is difficult. Autoregressive speech generation offers a promising solution, as demonstrated in models such as RichTTS~\cite{bai2024dmel} and VALL-T~\cite{du2024vall}.

Another significant challenge is that TTS models with text encoders struggle to handle streaming text input. These models typically need to re-encode the text sequence when new content arrives—a limitation affecting FastSpeech2's transformer encoder, Tacotron's LSTM encoder, and E3 TTS's BERT~\cite{koroteev2021bert} encoder. All these architectures face difficulty synthesizing natural speech with limited context.

Recent work~\cite{dekel2024speak} uses non-attentive Tacotron~\cite{shen2020non} for speech generation by distilling from a non-streaming TTS with limited access to future context. However, their architecture demonstrates limited zero-shot capability. Another work~\cite{dang2024zero} upgraded LiveSpeech~\cite{dang2024livespeech} from full-text audio synthesis to text-chunk synthesis. However, their model encounters misalignment issues between speech generation and text chunks. Although they introduce a CTC-ASR model to generate graphemes and guide chunk generation, this approach complicates the overall architecture and potentially introduces train-test mismatching issues.

Transducer-based TTS approaches~\cite{kim2023transduce} offer improved alignment design, but their application in dual-streaming settings remains under-explored. 

In this work, we propose SpeakStream, a decoder-only dual-streaming TTS model. By training a decoder-only transformer model with interleaved speech-text sequences, SpeakStream can store all generated chunks in its key-value cache, providing complete context without information loss. By predicting EOS tokens as chunk boundaries, our model avoids misalignment issues and eliminates the need for CTC aligners during inference. Furthermore, by controlling the interleaving window size, our model can attend to both current and future text chunks, ensuring minimal latency for the first audio chunk while maintaining smooth transitions for subsequent chunks.

Concurrently, \cite{yang2024interleaved} also found that decoder-only TTS models are suitable for dual-streaming synthesis. Unlike our model, they interleave text and speech with a fixed ratio rather than using alignment information, which could lead to complicated attention patterns when there are large variations in speaking rate. 
Furthermore, it might be difficult to precisely tie the streaming audio to the corresponding input text, which can be useful in interactive applications where a conversational agent is interruptible.
\section{Method}
Our approach enables streaming text-to-speech by introducing an interleaved representation of text and speech tokens in a decoder-only architecture.
The model processes these interleaved sequences to generate speech output with low latency. 
This section describes our model architecture, token representations, interleaving schemes, and inference method.
\subsection{Text and Speech Representation}

\noindent\textbf{Text Token Representation}: We use character-level embeddings to capture fine-grained linguistic features.
In this way, each word $w_t$ consists of $x$ character embeddings ${c_{t_1}, c_{t_2}, \cdots, c_{t_x}}$.

\noindent\textbf{Speech Token Representation}: We adopt the dMel~\cite{bai2024dmel} tokenization approach, which discretizes mel-filterbank channels into discrete intensity bins.
This simple yet effective discretization method preserves both acoustic and semantic information, making it ideal for our streaming synthesis task.
For each word $w_t$, its corresponding audio chunk $s_t$ consists of $y$ dMel embeddings ${f_{t_1}, f_{t_2}, \cdots, f_{t_y}}$.

\begin{figure}[ht!]
    \centering
    \includegraphics[width=0.65\linewidth]{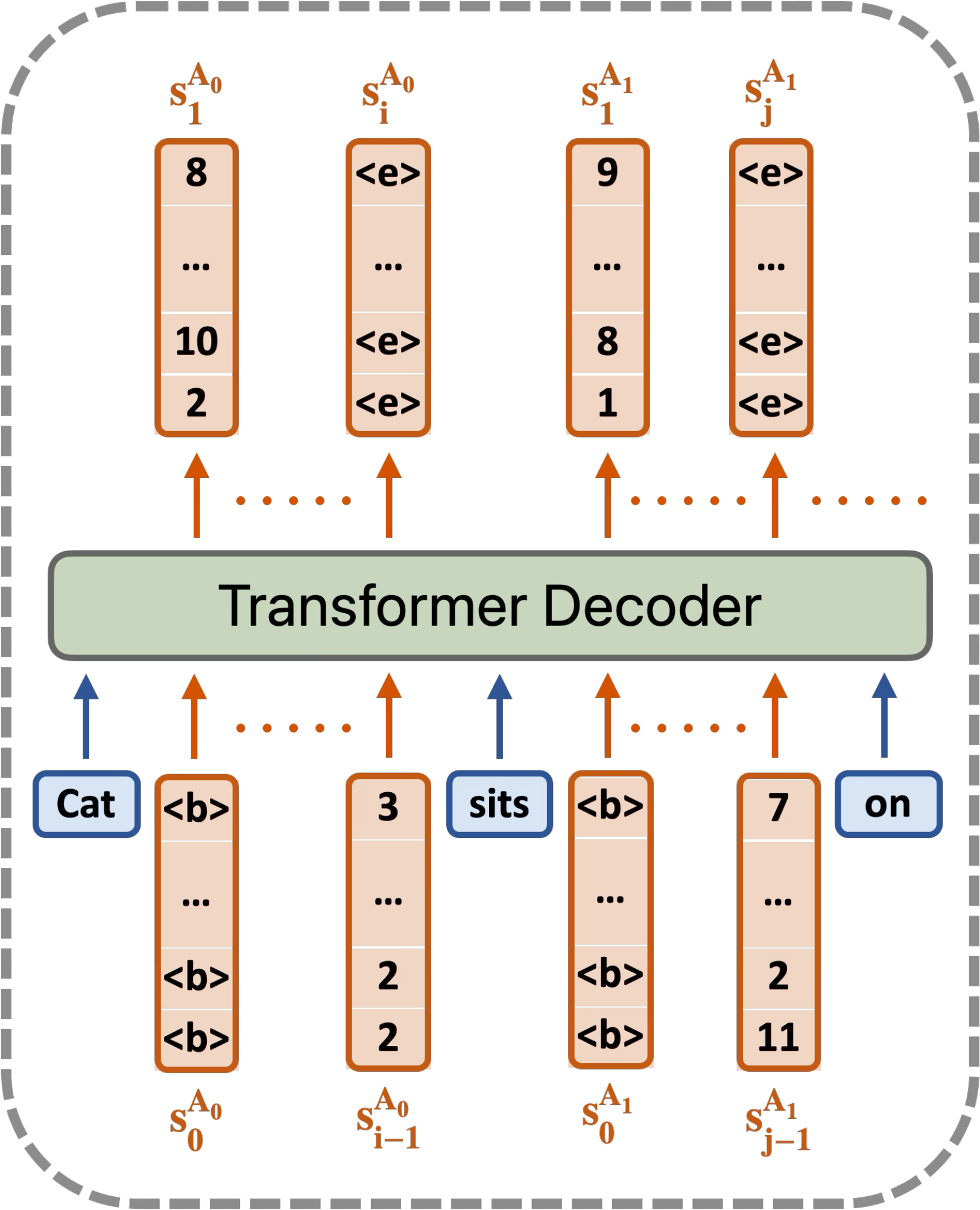}
    \caption{SpeakStream transformer decoder architecture.}
    \label{fig:main}
    \vspace{-0.4cm}
\end{figure}

\subsection{SpeakStream Model}
SpeakStream is built upon a vanilla transformer decoder architecture similar to RichTTS~\cite{bai2024dmel}.
Compared to RichTTS, the key difference is how SpeakStream constructs the transformer input sequence.
In RichTTS~\cite{bai2024dmel}, the input to the model is:
\begin{tcolorbox}[colback=white,center, width=0.9\linewidth,height=0.1\linewidth ]
\vspace{-0.5cm}
\begin{equation*}
[w_{\text{bos}}, w_1, \cdots, w_t, w_{\text{eos}}, s_{\text{bos}}, s_1, \cdots, s_t, s_{\text{eos}}]
\end{equation*}
\end{tcolorbox}
\noindent where $w_{\text{bos}}$ and $w_{\text{eos}}$ are the beginning and end text embeddings, $s_{\text{bos}}$ and $s_{\text{eos}}$ are the beginning and end speech embeddings. $t$ is the number of words.

SpeakStream, on the other hand interleaves the sequence above by inserting speech between text.
A simplified illustration of the input to SpeakStream is:
\begin{tcolorbox}[colback=white,center, width=0.8\linewidth, height=0.18\linewidth ]
    \vspace{-0.4cm}  
    \begin{align*}
    & [T_1, A_1, T_2, A_2, \cdots, T_x, A_x] \\
    & T_i = w_i, \quad A_i = s_{\text{bos}}, s_i, s_{\text{eos}} , \quad x = t 
    \end{align*}
\end{tcolorbox}
\noindent where text and speech are interleaved one by one.
To establish precise temporal correspondence between words and speech frames, we utilize the A3T's alignment mechanism~\cite{pmlr-v162-bai22d}.

By training a decoder-only transformer model on such interleaved sequences, SpeakStream learns to synthesize the current speech segment $A_i$ conditioned on the current text segment $T_i$, previous speech segments $A_{<i}$, and previous text segments $T_{<i}$.
Compared to existing streaming solutions that synthesize each text chunk independently, SpeakStream offers several advantages:
\begin{enumerate}
    \item Each speech segment $A_i$ is conditioned on the complete speech history $A_{<i}$, ensuring acoustic coherence and consistency;
    \item Each speech segment $A_i$ is conditioned on the complete text history $T_{<i}$, maintaining semantic precision;
    \item With access to comprehensive speech and text history, our model achieves high-quality streaming synthesis even with short text segments $T_i$, enabling lower latency;
    \item The model's kv-cache architecture efficiently maintains all historical context during inference, significantly reducing computational overhead and enabling fast generation;
    \item The model eliminates the need for a force aligner during inference by predicting the EOS token for each speech segment and processing text segments accordingly.
\end{enumerate}

\subsection{Interleaving Schemes}
There is a trade-off between latency and accuracy of streaming TTS.
To lower the latency, the length of each $T_i$ should be short.
However, given the existence of polyphonic words, the speech synthesis of certain words must consider not only the preceding words but also the subsequent words in the context sequence.
Therefore, $T_{<=i}$ should have additional words beyond $A_{i}$'s corresponding words to maintain synthesis accuracy.

To address this trade-off, we design two interleaving schemes, with each scheme having multiple variants by adjusting the \textbf{text window length} $m$ and \textbf{speech hop length} $n$, where $1\leq n\leq m$ and $m\geq 1$. This ensures the first $n$ words of $T_i$ correspond to $A_i$, while the remaining $(m-n)$ words provide future context.

\begin{tcolorbox}[colback=white,
    colframe=blue!60!black,
    arc=1mm,
    title={\textbf{Scheme 1}},
    fonttitle=\color{white}\small,
    coltitle=blue!60!black, height=0.6\linewidth ]
    \vspace{-0.4cm}
    \begin{align*}
        & [T_1, A_1, T_2, A_2, \cdots, T_x, A_x] \\
        & T_i = w_{n(i-1)+1},\cdots, w_{min(t, n(i-1)+m)} \\
        & A_i = s_{\text{bos}}, s_{n(i-1)+1},\cdots, s_{min(t, n\cdot i)}, s_{\text{eos}}, \quad x = \left\lceil\frac{t}{n}\right\rceil
        \end{align*}
    
    \textbf{Example} \text{ (}$m$=3, $n$=2, $t$=8\text{):} 
    \begin{align*}
        [\textcolor{TextColor}{w_1, w_2, w_3}, \textcolor{SpeechColor}{s_{\text{bos}}, s_1, s_2, s_{\text{eos}}}, \textcolor{TextColor}{w_3, w_4, w_5}, \textcolor{SpeechColor}{s_{\text{bos}}, s_3, s_4, s_{\text{eos}}},\\
        \textcolor{TextColor}{w_5, w_6, w_7}, \textcolor{SpeechColor}{s_{\text{bos}}, s_5, s_6, s_{\text{eos}}}, \textcolor{TextColor}{w_7, w_8}, \textcolor{SpeechColor}{s_{\text{bos}}, s_7, s_8, s_{\text{eos}}}]
    \end{align*}
\end{tcolorbox}
As shown in Scheme 1, we repeat text tokens to ensure the first $n$ words of $T_i$ correspond to $A_i$, while the remaining $(m-n)$ words provide future context.

\begin{tcolorbox}[colback=white,
    colframe=blue!60!black,
    arc=1mm,
    title={\textbf{Scheme 2}},
    fonttitle=\color{white}\small,
    coltitle=blue!60!black, height=0.7\linewidth ]
    \vspace{-0.4cm}
    \begin{align*}
    & [T_1, A_1, T_2, A_2, \cdots, T_x, A_x, A_{x+1}, \cdots, A_y] \\
    & T_i = \begin{cases}
        w_1,\cdots, w_{m} & i = 1\\
        w_{n(i-2)+m+1},\cdots, w_{min(t, n(i-1)+m)} & i >1 
    \end{cases} \\
    & A_i = s_{\text{bos}}, s_{n(i-1)+1},\cdots, s_{min(t, n\cdot i)}, s_{\text{eos}} &\\
    & x = \left\lceil\frac{t-m}{n}\right\rceil +1, \quad y = \left\lceil\frac{t}{n}\right\rceil 
\end{align*}
\textbf{Example} \text{ (}$m$=3, $n$=2, $t$=8\text{):}
\begin{align*}
    [\textcolor{TextColor}{w_1, w_2, w_3}, \textcolor{SpeechColor}{s_{\text{bos}}, s_1, s_2, s_{\text{eos}}}, \textcolor{TextColor}{w_4, w_5}, \textcolor{SpeechColor}{s_{\text{bos}}, s_3, s_4, s_{\text{eos}}},\\
    \textcolor{TextColor}{w_6, w_7}, \textcolor{SpeechColor}{s_{\text{bos}}, s_5, s_6, s_{\text{eos}}}, \textcolor{TextColor}{w_8}, \textcolor{SpeechColor}{s_{\text{bos}}, s_7, s_8, s_{\text{eos}}}]
\end{align*}
\end{tcolorbox}

Scheme 2 maintains the same $(m-n)$ contextual words but avoids token repetition, creating a more compact representation at the cost of more complex attention patterns.

\subsection{Streaming Inference}
During inference, our model enables true streaming generation through an autoregressive process that maintains the interleaved structure. Given a streaming text sequence, the model:
\begin{enumerate}
\item Waits until receiving the first segment of text words and generates their corresponding dMel tokens, which are converted into waveforms in tandem, using a streaming Mel-to-wave vocoder;
\item Uses both the generated speech features and the next segment of text words as context for generating the next speech segment by updating the kv-cache state;
\item Repeats this process until the entire text is synthesized.
\end{enumerate}

\noindent The simplicity of dMel tokenization allows our model to generate high-quality speech in a streaming fashion without the complexity of managing multiple token types or separate acoustic and semantic representations. The primary latency in our system comes from accumulating the first segment of text words, making the granularity parameter $m$ a direct control for the latency-quality trade-off.

\subsection{VocStream: Streaming Vocoder for Streaming TTS}

While enabling streaming TTS removes one bottleneck, the next major latency contributor in the system is the vocoder, which converts Mel-spectrograms into waveforms. In convolution-based vocoders~\cite{yamamoto2020parallel,kong2020hifi,leebigvgan,siuzdakvocos,shi2024non}, latency primarily arises from the need to accumulate several Mel frames to provide sufficient context for the convolutional stack to generate waveform chunks, in addition to the inherent model inference time.

To achieve end-to-end streaming synthesis, we propose VocStream, a fully streamable vocoder architecture. VocStream comprises two components: a streaming upsampler and a streaming vocoder, both built upon the ParallelWaveGAN~\cite{yamamoto2020parallel} framework. This two stage modeling is needed to preserve high quality reconstruction while having streaming property.

The VocStream upsampler is designed to increase both the temporal resolution and the feature dimensionality of the Mel-spectrogram---e.g., upsampling from 25ms to 6.25ms per frame and from 80 to 120 frequency channels. The VocStream vocoder then synthesizes the waveform from this higher-resolution Mel representation.

Crucially, both components rely exclusively on causal convolutional layers, ensuring that the system requires only a single Mel frame to produce the first waveform chunk. This design allows VocStream to operate with a fixed latency of one Mel frame, making it suitable for real-time, fully streamable speech synthesis.

\begin{table*}[!th]\centering
    \small
    \caption{WER of SpeakStream with Scheme 1~(S1) and Scheme 2~(S2) evaluated by WhisperX ASR~(base.en). \\ The WER of groudtruth audio is~2.09.}\label{tab:main}
    \begin{tabular}{l*{6}{cc|}c}\toprule
        \rowcolor{white}
    & \multicolumn{2}{c}{Hop n=1} & \multicolumn{2}{c}{Hop n=2} & \multicolumn{2}{c}{Hop n=3} & \multicolumn{2}{c}{Hop n=4} & \multicolumn{2}{c}{Hop n=5} & \multicolumn{2}{c}{Hop n=6} & $\infty$ \\\midrule
    \rowcolor{gray!15}
     XTTS-V2~\cite{coqui_tts_2021} & \multicolumn{2}{c}{222.28} & \multicolumn{2}{c}{45.01} & \multicolumn{2}{c}{30.92} & \multicolumn{2}{c}{28.97} & \multicolumn{2}{c}{17.13} & \multicolumn{2}{c}{15.28} & \textbf{3.67} \\
     \rowcolor{white}
    RichTTS~\cite{bai2024dmel} & \multicolumn{2}{c}{68.18} & \multicolumn{2}{c}{30.20} & \multicolumn{2}{c}{20.30} & \multicolumn{2}{c}{16.81} & \multicolumn{2}{c}{13.07} & \multicolumn{2}{c}{10.55} & \textbf{3.28} \\
    \rowcolor{gray!15}
    $n$-gram-RichTTS & \multicolumn{2}{c}{60.4} & \multicolumn{2}{c}{26.58} & \multicolumn{2}{c}{21.04} & \multicolumn{2}{c}{12.06} & \multicolumn{2}{c}{6.64} & \multicolumn{2}{c}{7.76} & \textbf{3.28} \\\midrule
    \rowcolor{white}
    SpeakStream & S1 & S2 & S1 & S2 & S1 & S2 & S1 & S2 & S1 & S2 & S1 & S2 & \\
    \rowcolor{gray!15}
    Window m=1 & \multicolumn{2}{c|}{7.47} &\multicolumn{2}{c|}{-} &\multicolumn{2}{c|}{-} & \multicolumn{2}{c|}{-} & \multicolumn{2}{c|}{-} &\multicolumn{2}{c|}{-} & - \\
    \rowcolor{white}
    Window m=2 & \textbf{4.50} & 5.26 & \multicolumn{2}{c|}{7.18} &\multicolumn{2}{c|}{-} & \multicolumn{2}{c|}{-} & \multicolumn{2}{c|}{-} &\multicolumn{2}{c|}{-} & - \\
    \rowcolor{gray!15}
    Window m=3 & \textbf{3.99} & 6.88 & \textbf{4.19} & 5.11 & \multicolumn{2}{c|}{4.78} & \multicolumn{2}{c|}{-} & \multicolumn{2}{c|}{-} &\multicolumn{2}{c|}{-} & - \\
    \rowcolor{white}
    Window m=4 & \textbf{3.88} & 6.16 & \textbf{3.80} & 5.09 & \textbf{4.36} & 5.35 & \multicolumn{2}{c|}{5.21} & \multicolumn{2}{c|}{-} &\multicolumn{2}{c|}{-} & - \\
    \rowcolor{gray!15}
    Window m=5 & \textbf{3.38} & 5.59 & \textbf{3.61} & 6.09 & \textbf{3.65} & 4.82 & 4.73 & \textbf{4.59} & \multicolumn{2}{c|}{4.52} &\multicolumn{2}{c|}{-} & - \\
    \rowcolor{white}
    Window m=6 & \textbf{4.26} & 6.20 & \textbf{3.61} & 4.36 & \textbf{3.93} & 4.52 & \textbf{4.36} & 6.14 & \textbf{4.86} & 4.95 & \multicolumn{2}{c|}{4.30} & - \\\bottomrule
    \end{tabular}
    \vspace{-0.4cm}
\end{table*}

\begin{table}[h!]
    \small
    \center
    \caption{Human evaluation results for synthesized speech naturalness and coherence~(95\% confidence interval).
    }\label{tab:human_eval}
    \resizebox{0.46\textwidth}{!}{
    \begin{tabular}{lccc}
        \toprule
        & NonStreaming & Streaming ($m$=4) & Streaming ($m$=6) \\
        \midrule
        & \multicolumn{3}{c}{\textbf{Naturalness}} \\
        \rowcolor{gray!15}
        GroundTruth & $\textbf{4.4}\pm0.1$ & - & - \\
        \rowcolor{white}
        RichTTS & $3.8\pm0.1$ & $2.2\pm0.1$ & $2.5\pm0.1$ \\
        \rowcolor{gray!15}
        XTTS & $3.9\pm0.1$ & $2.1\pm0.1$ & $2.5\pm0.1$ \\
        \rowcolor{white}
        SpeakStream & - & $\textbf{3.7}\pm0.1$ & $\textbf{3.5}\pm0.1$ \\
        \midrule
        & \multicolumn{3}{c}{\textbf{Coherence}} \\
        \rowcolor{gray!15}
        GroundTruth & $\textbf{4.2}\pm0.0$ & - & - \\
        \rowcolor{white}
        RichTTS & $3.9\pm0.1$ & $2.3\pm0.1$ & $2.2\pm0.1$ \\
        \rowcolor{gray!15}
        XTTS & $4.1\pm0.1$ & $1.8\pm0.1$ & $2.7\pm0.1$ \\
        \rowcolor{white}
        SpeakStream & - & $\textbf{3.9}\pm0.1$ & $\textbf{3.8}\pm0.1$ \\
        \bottomrule
    \end{tabular}
    }
    \vspace{-0.2cm}
\end{table}

\begin{figure*}[ht]
    \vspace{0.5cm}
    \centering
    \includegraphics[width=0.9\linewidth]{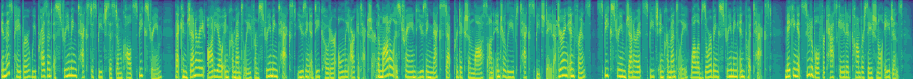}
    \caption{A 30-second Mel-spectrogram generated from the SpeakStream model. It corresponds to the audio in our demo, whose transcription is the abstract section of this paper.}
    \label{fig:spectrogram}
    \vspace{-0.2cm}
\end{figure*}

\begin{table}[!th]\centering
    \caption{Latency~(ms) of SpeakStream with ParallelWaveGAN (PWG,~1M) or VocStream (11.9M) vocoders, tested with Apple Silicon (Mac mini, M4 Pro, 64GB, 2024). \\Player latency is less than 0.2ms.
    }\label{tab:latency}
    \resizebox{0.46\textwidth}{!}{
    \begin{tabular}{ccccc}\toprule
    \rowcolor{white}
    SpeakStream  & Vocoder & TTS (ms) & Vocoder (ms)  &Total (ms) \\
    \midrule
    Window m=3 & \multirow{3}{*}{PWG} & 28$_{\pm 2}$ & 152$_{\pm 12}$ & 180$_{\pm 14}$\\
    Window m=4 &  & 33$_{\pm 8}$ & 150$_{\pm 5}$ & 183$_{\pm 13}$\\
    Window m=5 &  & 34$_{\pm 9}$ & 150$_{\pm 6}$ & 184$_{\pm 15}$  \\
    \midrule
    Window m=3 & \multirow{3}{*}{VocStream} & 27$_{\pm 2}$ & 13$_{\pm 1}$ & 40$_{\pm 3}$ \\
    Window m=4 &  & 28$_{\pm 2}$ & 13$_{\pm 1}$ & 41$_{\pm 4}$ \\
    Window m=5 &  & 29$_{\pm 1}$ & 13$_{\pm 1}$ & 45$_{\pm 15}$ \\
    \bottomrule
    \end{tabular}
    }
    \vspace{-0.2cm}
\end{table}

\section{Experiments}
\subsection{Setup}
We conduct experiments using the LJSpeech~\cite{ljspeech17} dataset, which consists of single-speaker English audio recordings at 22kHz with read speech from LibriVox.
The reason we choose LJSpeech is that it has been widely used in single-speaker setup and is suited for our experiments on latency and accuracy.
Following RichTTS~\cite{bai2024dmel}, our model comprises 36 layers of transformer decoder with 258M parameters.
The dMel feature is exactly the same as \cite{bai2024dmel} with 25ms hop length, 80 channels, 16 bins, and ParallelWaveGAN vocoder~\cite{yamamoto2020parallel}.
The training configuration is the same as RichTTS: we train all models using the Adam optimizer with a learning rate of 1e-3, warmup of 5k steps, cosine learning rate schedule, and gradient clipping of 1.0. Each model is trained for 100k steps with a batch size of 1 hour on A100 machines using mixed precision BF16.

The baseline models are RichTTS~\cite{bai2024dmel} and XTTS~\cite{coqui_tts_2021}, which are trained on complete text and audio pairs.
We choose XTTS as the baseline because it is the best open-source TTS model with streaming inference capability.
We also train multiple $n$-gram versions of RichTTS, where the models are trained with short segments.
For TTS evaluation, we utilize WhisperX~(``base.en'')~\cite{bain2022whisperx,radford2023robust} to transcribe our generated speech into text and calculate the Word Error Rate (WER).

For fully streaming text-to-speech experiments, we train a streaming vocoder (VocStream) with LibriTTS-R dataset~\cite{koizumi2023libritts}.
It employs a generator comprising 30 WaveNet residual blocks (4.6M parameters) and a discriminator with 10 layers. 
The VocStream vocoder takes the 120-channel, 6.25ms-resolution Mel features from the upsampler and synthesizes a 24kHz waveform. 
It consists of a 44-layer WaveNet-based generator (7.3M parameters) and a 10-layer discriminator.


\subsection{Main Results}
Our main results are presented in Table~\ref{tab:main}.
The experiments show that directly applying RichTTS to streaming segments significantly degrades generation quality, with WER exceeding 68\% for unigram word synthesis.
XTTS performs even worse, with WER exceeding 222\% for unigram word synthesis.
Upon investigating its generation, we find that XTTS hallucinates words and phonemes, leading to high insertion errors.
Although these two models perform well in full text synthesis, with WER below 4\%, they are not suitable for streaming synthesis.
For RichTTS trained with short segments, the WER is significantly reduced, but it still remains above 60\% for unigram word synthesis.
In contrast, SpeakStream achieves WER around 7\% for unigram word synthesis, and below 5\% when more context is provided.
Importantly, \textbf{SpeakStream's performance is comparable to RichTTS's full text synthesis when ${m=5}$ and ${n=1}$. At the same time, SpeakStream has low latency (5 words) while RichTTS has high latency (full text).}
This result demonstrates that SpeakStream can achieve high-quality streaming synthesis with interleaved text and speech inputs.

For SpeakStream, Scheme 1~(S1) interleaving consistently outperforms Scheme 2~(S2). This performance difference stems from the varying speech durations of each $A_i$, which impacts S2 more substantially than S1. With S1, the model can easily locate the $n$ corresponding words in $T_i$ adjacent to~$A_i$, while the remaining $(m-n)$ words provide supplementary pronunciation context. In contrast, S2 requires more complex attention patterns as $A_i$'s corresponding words are separated by variable-length gaps determined by $A_{i-1}$'s duration. Based on these findings, we adopt S1 as SpeakStream's interleaving scheme for subsequent analysis.

The results reveal that configurations where $m=n$ yield higher WER, indicating that additional text tokens enhance segment synthesis quality. Performance improves notably when $(m-n) > 1$. As expected, increasing $m$ generally improves performance by providing richer context. The optimal configuration occurs at $m=5, n=1$, achieving 3.38 WER, comparable to RichTTS's non-streaming (high latency) performance. However, performance deteriorates beyond $m=5$ due to the $(m-n)$ word repetition creating excessively long text sequences, suggesting that overly large context windows don't necessarily improve accuracy for SpeakStream.

\subsection{Human Evaluation}
We conducted a human evaluation to assess the quality of synthesized speech from different streaming models.
We randomly sampled 100 segments from the LJSpeech dev set and asked human evaluators to rate the naturalness and coherence of each segment on a scale of 1 to 5.
Each segment was evaluated by 7 random annotators, with 284 unique annotators in total. 
95\% confidence interval is reported.
We evaluated 4-gram and 6-gram versions of RichTTS and XTTS, corresponding to $(m=4, n=2)$ and $(m=6, n=2)$ configurations of SpeakStream.
The results are shown in Table~\ref{tab:human_eval}.

We observed that non-streaming TTS results are similar between RichTTS and XTTS, with XTTS slightly outperforming RichTTS in both naturalness and coherence.
However, when applied in streaming settings, XTTS's performance drops significantly, especially in coherence.
For streaming models, human evaluators rated SpeakStream as coherent as non-streaming RichTTS, suggesting that SpeakStream successfully maintains coherence despite the model's streaming nature.
To visualize the coherence of SpeakStream, here we show a 30-second spectrogram in Figure~\ref{fig:spectrogram}.
This figure shows that it is hard to identify where SpeakStream predicts the EOS of speech.

\subsection{Latency Analysis}
We conducted latency analysis of SpeakStream with streaming text input and streaming audio output. Our implementation features sequential word input to the TTS model, while the streaming output pipeline consists of streaming frame generation from TTS, streaming waveform synthesis from the vocoder, and real-time audio player.
We implemented the complete system using MLX~\cite{mlx2023} and deployed it on a Mac Mini 2024 with M4 Pro Apple Silicon (64GB RAM). For evaluation, we randomly sampled 25 sentences
from the LibriSpeech~\cite{panayotov2015librispeech} dev-clean set and measured the performance of SpeakStream models trained with configuration S1, n=1. 
We report three latency metrics:
\begin{enumerate}
    \item \textbf{Total latency}: Time elapsed between the TTS model receiving its first word and the audio player outputting the first waveform chunk.
    \item \textbf{Vocoder latency}: Time elapsed between the vocoder receiving its first frame input and generating the first chunk of waveform.
    \item \textbf{TTS latency}: Time elapsed between the TTS model receiving its first word and generating its first frame.
\end{enumerate}
\textbf{\textit{Note, we are interested not in time when the first waveform outputted but rather in time of the first spoken phoneme.\footnote{In \href{https://github.com/Picovoice/tts-latency-benchmark}{tts-latency-benchmark} authors measured the first-token to speech considering first byte of speech. However, the model can learn to delay the actual phoneme generation due to various factors, thus increasing the real latency. That is why we are focusing on measuring first-token to actual phoneme as the latency of TTS-vocoder-player system.}}} During inference we prompt our model with small silence audio (200-400ms), so that model start generation of actual phoneme right away, see Fig.~\ref{fig:spectrogram}. \textit{For all samples of latency benchmark we observe the first generated dMel-frame containing a speech pattern thus SpeakStream has no extra latency delay for the first spoken phoneme to be produced.}
It should be noted that this paper primarily focuses on reducing TTS latency. In real conversational agents, the text token generation time should also be considered; however, since this depends on the LLM's size and inference infrastructure, we use word count rather than milliseconds to measure this component.

Our results in Table~\ref{tab:latency} demonstrate that SpeakStream achieves low latency, requiring only around 30ms to generate the first frame.
Also, SpeakStream is only waiting m words before it starts the generation process.
This efficiency remains consistent across all configurations. 
Non-streaming ParallelWaveGAN~\cite{yamamoto2020parallel} requires 10 frames to be available before producing output waveform: having inference speed of TTS around 15ms per frame, we get vocoder latency to be around 150ms.
Switching to our streaming VocStream vocoder (needs only 1 frame to start generation), we get latency of 13ms.
The total response time of SpeakStream combined with our VocStream vocoder is below 50ms, which is 2.4x faster than favorable latency of 120ms for interactive applications (see e.g.~\cite{moshi} and \href{https://github.com/Picovoice/tts-latency-benchmark}{tts-latency-benchmark}).

\section{Conclusion}

We presented SpeakStream, a decoder-only streaming TTS system that enables real-time speech synthesis through interleaved text-speech modeling. 
Our extensive experiments demonstrate that SpeakStream successfully bridges the quality gap between streaming and non-streaming TTS systems, achieving WER comparable to full-text synthesis while operating in a streaming fashion with low latency. The system's ability to maintain coherence across segments, as confirmed by human evaluations, makes it a promising solution for interactive applications where both responsiveness and naturalness are critical.
\textit{System of SpeakStream, streaming vocoder and audio player achieves less than 50ms first-token to speech latency on Mac Mini with M4 Pro and 64GB.}
Future work could explore extending this approach to multi-speaker settings, larger datasets, and cross-lingual applications.

\newpage
\bibliographystyle{IEEEtran}
\bibliography{mybib}

\begin{thebibliography}{10}
\providecommand{\url}[1]{#1}
\csname url@samestyle\endcsname
\providecommand{\newblock}{\relax}
\providecommand{\bibinfo}[2]{#2}
\providecommand{\BIBentrySTDinterwordspacing}{\spaceskip=0pt\relax}
\providecommand{\BIBentryALTinterwordstretchfactor}{4}
\providecommand{\BIBentryALTinterwordspacing}{\spaceskip=\fontdimen2\font plus
\BIBentryALTinterwordstretchfactor\fontdimen3\font minus \fontdimen4\font\relax}
\providecommand{\BIBforeignlanguage}[2]{{%
\expandafter\ifx\csname l@#1\endcsname\relax
\typeout{** WARNING: IEEEtran.bst: No hyphenation pattern has been}%
\typeout{** loaded for the language `#1'. Using the pattern for}%
\typeout{** the default language instead.}%
\else
\language=\csname l@#1\endcsname
\fi
#2}}
\providecommand{\BIBdecl}{\relax}
\BIBdecl

\bibitem{borsos2023audiolm}
Z.~Borsos, R.~Marinier, D.~Vincent, E.~Kharitonov, O.~Pietquin, M.~Sharifi, D.~Roblek, O.~Teboul, D.~Grangier, M.~Tagliasacchi \emph{et~al.}, ``Audiolm: a language modeling approach to audio generation,'' \emph{IEEE/ACM transactions on audio, speech, and language processing}, vol.~31, pp. 2523--2533, 2023.

\bibitem{defossez2024moshi}
A.~D{\'e}fossez, L.~Mazar{\'e}, M.~Orsini, A.~Royer, P.~P{\'e}rez, H.~J{\'e}gou, E.~Grave, and N.~Zeghidour, ``Moshi: a speech-text foundation model for real-time dialogue,'' \emph{arXiv preprint arXiv:2410.00037}, 2024.

\bibitem{Qwen2.5-Omni}
J.~Xu, Z.~Guo, J.~He, H.~Hu, T.~He, S.~Bai, K.~Chen, J.~Wang, Y.~Fan, K.~Dang, B.~Zhang, X.~Wang, Y.~Chu, and J.~Lin, ``Qwen2.5-omni technical report,'' \emph{arXiv preprint arXiv:2503.20215}, 2025.

\bibitem{nguyen2025spirit}
T.~A. Nguyen, B.~Muller, B.~Yu, M.~R. Costa-Jussa, M.~Elbayad, S.~Popuri, C.~Ropers, P.-A. Duquenne, R.~Algayres, R.~Mavlyutov \emph{et~al.}, ``Spirit-lm: Interleaved spoken and written language model,'' \emph{Transactions of the Association for Computational Linguistics}, vol.~13, pp. 30--52, 2025.

\bibitem{sakshi2024mmau}
S.~Sakshi, U.~Tyagi, S.~Kumar, A.~Seth, R.~Selvakumar, O.~Nieto, R.~Duraiswami, S.~Ghosh, and D.~Manocha, ``Mmau: A massive multi-task audio understanding and reasoning benchmark,'' \emph{arXiv preprint arXiv:2410.19168}, 2024.

\bibitem{dang2024zero}
T.~Dang, D.~Aponte, D.~Tran, T.~Chen, and K.~Koishida, ``Zero-shot text-to-speech from continuous text streams,'' \emph{arXiv preprint arXiv:2410.00767}, 2024.

\bibitem{dekel2024speak}
A.~Dekel, S.~Shechtman, R.~Fernandez, D.~Haws, Z.~Kons, and R.~Hoory, ``Speak while you think: Streaming speech synthesis during text generation,'' in \emph{ICASSP 2024-2024 IEEE International Conference on Acoustics, Speech and Signal Processing (ICASSP)}.\hskip 1em plus 0.5em minus 0.4em\relax IEEE, 2024, pp. 11\,931--11\,935.

\bibitem{yang2024interleaved}
Y.~Yang, Z.~Ma, S.~Liu, J.~Li, H.~Wang, L.~Meng, H.~Sun, Y.~Liang, R.~Xu, Y.~Hu \emph{et~al.}, ``Interleaved speech-text language models are simple streaming text to speech synthesizers,'' \emph{arXiv preprint arXiv:2412.16102}, 2024.

\bibitem{bai2024dmel}
H.~Bai, T.~Likhomanenko, R.~Zhang, Z.~Gu, Z.~Aldeneh, and N.~Jaitly, ``dmel: Speech tokenization made simple,'' \emph{arXiv preprint arXiv:2407.15835}, 2024.

\bibitem{pmlr-v162-bai22d}
\BIBentryALTinterwordspacing
H.~Bai, R.~Zheng, J.~Chen, M.~Ma, X.~Li, and L.~Huang, ``{A}$^3${T}: Alignment-aware acoustic and text pretraining for speech synthesis and editing,'' in \emph{Proceedings of the 39th International Conference on Machine Learning}, ser. Proceedings of Machine Learning Research, K.~Chaudhuri, S.~Jegelka, L.~Song, C.~Szepesvari, G.~Niu, and S.~Sabato, Eds., vol. 162.\hskip 1em plus 0.5em minus 0.4em\relax PMLR, 17--23 Jul 2022, pp. 1399--1411. [Online]. Available: \url{https://proceedings.mlr.press/v162/bai22d.html}
\BIBentrySTDinterwordspacing

\bibitem{casanova2022yourtts}
E.~Casanova, J.~Weber, C.~D. Shulby, A.~C. Junior, E.~G{\"o}lge, and M.~A. Ponti, ``Yourtts: Towards zero-shot multi-speaker tts and zero-shot voice conversion for everyone,'' in \emph{International Conference on Machine Learning}.\hskip 1em plus 0.5em minus 0.4em\relax PMLR, 2022, pp. 2709--2720.

\bibitem{du2024cosyvoice}
Z.~Du, Q.~Chen, S.~Zhang, K.~Hu, H.~Lu, Y.~Yang, H.~Hu, S.~Zheng, Y.~Gu, Z.~Ma \emph{et~al.}, ``Cosyvoice: A scalable multilingual zero-shot text-to-speech synthesizer based on supervised semantic tokens,'' \emph{arXiv preprint arXiv:2407.05407}, 2024.

\bibitem{gao2023e3}
Y.~Gao, N.~Morioka, Y.~Zhang, and N.~Chen, ``E3 tts: Easy end-to-end diffusion-based text to speech,'' in \emph{2023 IEEE Automatic Speech Recognition and Understanding Workshop (ASRU)}.\hskip 1em plus 0.5em minus 0.4em\relax IEEE, 2023, pp. 1--8.

\bibitem{ren2020fastspeech}
Y.~Ren, C.~Hu, X.~Tan, T.~Qin, S.~Zhao, Z.~Zhao, and T.-Y. Liu, ``Fastspeech 2: Fast and high-quality end-to-end text to speech,'' \emph{arXiv preprint arXiv:2006.04558}, 2020.

\bibitem{wang2017tacotron}
Y.~Wang, R.~Skerry-Ryan, D.~Stanton, Y.~Wu, R.~J. Weiss, N.~Jaitly, Z.~Yang, Y.~Xiao, Z.~Chen, S.~Bengio \emph{et~al.}, ``Tacotron: Towards end-to-end speech synthesis,'' \emph{arXiv preprint arXiv:1703.10135}, 2017.

\bibitem{openai_tts}
\BIBentryALTinterwordspacing
{OpenAI}. (2024) Text-to-speech guide. [Online]. Available: \url{https://platform.openai.com/docs/guides/text-to-speech}
\BIBentrySTDinterwordspacing

\bibitem{10096566}
G.~Shopov, S.~Gerdjikov, and S.~Mihov, ``Streamspeech: Low-latency neural architecture for high-quality on-device speech synthesis,'' in \emph{ICASSP 2023 - 2023 IEEE International Conference on Acoustics, Speech and Signal Processing (ICASSP)}, 2023, pp. 1--5.

\bibitem{mlx2023}
\BIBentryALTinterwordspacing
A.~Hannun, J.~Digani, A.~Katharopoulos, and R.~Collobert, ``{MLX}: Efficient and flexible machine learning on apple silicon,'' 2023. [Online]. Available: \url{https://github.com/ml-explore}
\BIBentrySTDinterwordspacing

\bibitem{du2024vall}
C.~Du, Y.~Guo, H.~Wang, Y.~Yang, Z.~Niu, S.~Wang, H.~Zhang, X.~Chen, and K.~Yu, ``Vall-t: Decoder-only generative transducer for robust and decoding-controllable text-to-speech,'' \emph{arXiv preprint arXiv:2401.14321}, 2024.

\bibitem{koroteev2021bert}
M.~V. Koroteev, ``Bert: a review of applications in natural language processing and understanding,'' \emph{arXiv preprint arXiv:2103.11943}, 2021.

\bibitem{shen2020non}
J.~Shen, Y.~Jia, M.~Chrzanowski, Y.~Zhang, I.~Elias, H.~Zen, and Y.~Wu, ``Non-attentive tacotron: Robust and controllable neural tts synthesis including unsupervised duration modeling,'' \emph{arXiv preprint arXiv:2010.04301}, 2020.

\bibitem{dang2024livespeech}
T.~Dang, D.~Aponte, D.~Tran, and K.~Koishida, ``Livespeech: Low-latency zero-shot text-to-speech via autoregressive modeling of audio discrete codes,'' \emph{arXiv preprint arXiv:2406.02897}, 2024.

\bibitem{kim2023transduce}
M.~Kim, M.~Jeong, B.~J. Choi, D.~Lee, and N.~S. Kim, ``Transduce and speak: Neural transducer for text-to-speech with semantic token prediction,'' in \emph{2023 IEEE Automatic Speech Recognition and Understanding Workshop (ASRU)}.\hskip 1em plus 0.5em minus 0.4em\relax IEEE, 2023, pp. 1--7.

\bibitem{yamamoto2020parallel}
R.~Yamamoto, E.~Song, and J.-M. Kim, ``Parallel wavegan: A fast waveform generation model based on generative adversarial networks with multi-resolution spectrogram,'' in \emph{ICASSP 2020-2020 IEEE International Conference on Acoustics, Speech and Signal Processing (ICASSP)}.\hskip 1em plus 0.5em minus 0.4em\relax IEEE, 2020, pp. 6199--6203.

\bibitem{kong2020hifi}
J.~Kong, J.~Kim, and J.~Bae, ``Hifi-gan: Generative adversarial networks for efficient and high fidelity speech synthesis,'' \emph{Advances in neural information processing systems}, vol.~33, pp. 17\,022--17\,033, 2020.

\bibitem{leebigvgan}
S.-g. Lee, W.~Ping, B.~Ginsburg, B.~Catanzaro, and S.~Yoon, ``Bigvgan: A universal neural vocoder with large-scale training,'' in \emph{The Eleventh International Conference on Learning Representations}.

\bibitem{siuzdakvocos}
H.~Siuzdak, ``Vocos: Closing the gap between time-domain and fourier-based neural vocoders for high-quality audio synthesis,'' in \emph{The Twelfth International Conference on Learning Representations}.

\bibitem{shi2024non}
R.~Shi, A.~B{\"a}r, M.~Sach, W.~Tirry, and T.~Fingscheidt, ``Non-causal to causal ssl-supported transfer learning: Towards a high-performance low-latency speech vocoder,'' in \emph{2024 18th International Workshop on Acoustic Signal Enhancement (IWAENC)}.\hskip 1em plus 0.5em minus 0.4em\relax IEEE, 2024, pp. 359--363.

\bibitem{coqui_tts_2021}
\BIBentryALTinterwordspacing
E.~G{\"o}lge and {The Coqui TTS Team}, ``Coqui {TTS}: A deep learning toolkit for {Text-to-Speech}, battle-tested in research and production,'' 1 2021. [Online]. Available: \url{https://www.coqui.ai}
\BIBentrySTDinterwordspacing

\bibitem{ljspeech17}
K.~Ito and L.~Johnson, ``The lj speech dataset,'' \url{https://keithito.com/LJ-Speech-Dataset/}, 2017.

\bibitem{bain2022whisperx}
M.~Bain, J.~Huh, T.~Han, and A.~Zisserman, ``Whisperx: Time-accurate speech transcription of long-form audio,'' \emph{INTERSPEECH 2023}, 2023.

\bibitem{radford2023robust}
A.~Radford, J.~W. Kim, T.~Xu, G.~Brockman, C.~McLeavey, and I.~Sutskever, ``Robust speech recognition via large-scale weak supervision,'' in \emph{International Conference on Machine Learning}.\hskip 1em plus 0.5em minus 0.4em\relax PMLR, 2023, pp. 28\,492--28\,518.

\bibitem{koizumi2023libritts}
Y.~Koizumi, H.~Zen, S.~Karita, Y.~Ding, K.~Yatabe, N.~Morioka, M.~Bacchiani, Y.~Zhang, W.~Han, and A.~Bapna, ``Libritts-r: A restored multi-speaker text-to-speech corpus,'' in \emph{Proc. Interspeech 2023}, 2023, pp. 5496--5500.

\bibitem{panayotov2015librispeech}
V.~Panayotov, G.~Chen, D.~Povey, and S.~Khudanpur, ``Librispeech: an asr corpus based on public domain audio books,'' in \emph{2015 IEEE international conference on acoustics, speech and signal processing (ICASSP)}.\hskip 1em plus 0.5em minus 0.4em\relax IEEE, 2015, pp. 5206--5210.

\bibitem{moshi}
\BIBentryALTinterwordspacing
A.~Défossez, L.~Mazaré, M.~Orsini, A.~Royer, P.~Pérez, H.~Jégou, E.~Grave, and N.~Zeghidour, ``Moshi: a speech-text foundation model for real-time dialogue,'' 2024. [Online]. Available: \url{https://arxiv.org/abs/2410.00037}
\BIBentrySTDinterwordspacing

\end{thebibliography}

\end{document}